\documentclass[preprint,12pt]{elsarticle}
\usepackage{amsfonts}  
\usepackage{amsmath}
\usepackage{xcolor}
\usepackage{amsthm}
\usepackage{multirow}
\usepackage{booktabs}
\usepackage{graphicx}
\usepackage{subcaption}
\usepackage{enumitem}
\usepackage{float}
\usepackage{url} 
\usepackage{longtable}
\usepackage{multirow}
\usepackage{caption}
\usepackage{hyperref}
\begin{document}

\begin{frontmatter}

\title{HybridoNet-Adapt: A Domain-Adapted Framework for Accurate Lithium-Ion Battery RUL Prediction}

\author[1]{Khoa~Tran}

\author[1]{Bao~Huynh}

\author[1]{Tri~Le}

\author[2]{Lam~Pham}

\author[3]{Vy-Rin~Nguyen}

\author[4]{Hung-Cuong Trinh\corref{cor1}}

\author[5,6]{Duong Tran Anh}

\cortext[cor1]{Corresponding author: \href{mailto:trinhhungcuong@tdtu.edu.vn}{trinhhungcuong@tdtu.edu.vn}}

\affiliation[1]{organization={AIWARE Limited Company},  
addressline={17 Huynh Man Dat Street, Hoa Cuong Bac Ward, Hai Chau District},  
city={Da Nang},  
postcode={550000},  
country={Vietnam}}  

\affiliation[2]{organization={AIT Austrian Institute of Technology GmbH},  
addressline={Giefinggasse 4},  
city={Vienna},  
postcode={1210},  
country={Austria}}  

\affiliation[3]{organization={Software Engineering Department, FPT University},  
city={Da Nang},  
postcode={550000},  
country={Vietnam}}  

\affiliation[4]{organization={Faculty of Information Technology, Ton Duc Thang University},  
city={Ho Chi Minh City},  
postcode={700000},  
country={Vietnam}}

\affiliation[5]{organization={Laboratory of Environmental Sciences and Climate Change, Institute for Computational Science and Artificial Intelligence, Van Lang University},  
city={Ho Chi Minh City},  
postcode={700000},  
country={Vietnam}}  

\affiliation[6]{organization={Faculty of Environment, School of Technology, Van Lang University},  
city={Ho Chi Minh City},  
postcode={700000},  
country={Vietnam}}

\begin{abstract}
Accurate prediction of the Remaining Useful Life (RUL) in Lithium-ion battery (LIB) health management systems is essential for ensuring operational reliability and safety. However, many existing methods assume that training and testing data follow the same distribution, limiting their ability to learn domain-invariant features and generalize effectively to unseen target domains. To approach this opportunity, we propose a novel RUL prediction framework and incorporates an domain adaptation (DA) technique. Our framework integrates a comprehensive signal preprocessing pipeline—including noise-reduction, feature extraction, and normalization—with a robust deep learning model named \textit{HybridoNet-Adapt}. The model comprises a feature extraction block built from Long Short-Term Memory (LSTM), Multihead Attention, and Neural Ordinary Differential Equation (NODE) layers, followed by two predictor modules implemented using linear layers. These predictors are balanced using trainable trade-off parameters. To enhance generalization to unseen data distributions, we introduce a domain adaptation strategy inspired by the Domain-Adversarial Neural Network (DANN) framework, replacing adversarial loss with Maximum Mean Discrepancy (MMD) to learn domain-invariant features. This approach enables effective transfer learning from source to target domains using cycling data. Experimental results demonstrate that our proposed method significantly outperforms traditional machine learning models such as XGBoost and Elastic Net, as well as deep learning baselines like Dual-input DNN, in predicting RUL. These findings highlight the potential of \textit{HybridoNet-Adapt} for reliable and scalable Battery Health Management (BHM).

\end{abstract}

\end{frontmatter}

\section{Introduction}
\subsection{Motivations}
Lithium-ion batteries (LIBs) \cite{goodenough2013li}, renowned for their affordability and high energy density, are extensively utilized \cite{guo2024managing, dunn2011electrical, zhang2023patent, larcher2015towards} in electric vehicles (EVs), portable devices, and energy storage stations. The global lithium-ion battery (LIB) market is projected to surpass 170 billion dollars by 2030 \cite{pillot2019rechargeable}. With the wide-ranging adoption of LIBs, interest in battery health management (BHM) has surged within both academia and industry in recent years. In a BHM system, several common and essential techniques are employed, including thermal management \cite{karimi2013thermal, zhang2023research}, fault diagnosis/detection \cite{du2024exploring}, state of charge (SOC) \cite{ma2021state, awadallah2016accuracy, hong2020online} and state of health (SOH) \cite{luo2022review} estimation, remaining useful life (RUL) prediction \cite{li2023development, ma2024accurate}, and cycle life early prediction \cite{severson2019data, ibraheem2023capacity, lin2021early, zhao2022lithium}. Among these, RUL prediction plays a crucial role in ensuring the reliability, safety, and optimal performance of LIBs over their lifespan. RUL can be assessed based on a capacity metric, such as SOH, or the number of remaining cycles the battery can undergo before reaching its end of life. Accurate RUL prediction allows for proactive maintenance, minimizing downtime, and enhancing the operational efficiency of LIBs. RUL prediction falls into three categories: model-based, data-driven, and hybrid approaches.

Traditional model-based approaches often utilize physics-based degradation models, such as the Double Exponential Model (DEM) \cite{ma2023two}, two-phase degradation models \cite{wang2023lithium}, and Markov Models \cite{zhang2023joint}, constructed using early-cycle data (200-500 cycles). These models aim to forecast the entire battery's capacity degradation curve. However, relying solely on maximum discharge capacity degradation during early cycles often leads to inaccuracies due to the influence of various factors (current, voltage, temperature, time) and sudden changes in degradation trends \cite{ma2023two, wang2023lithium}.

Data-driven models predict the RUL of LIBs by analyzing data during current cycles. Techniques like dual-input Deep Neural Networks (DNN) \cite{xia2023historical}, 1D Convolutional Neural Networks (1DCNN) \cite{kiranyaz20211d}, Dense layers \cite{javid2021relu}, Long Short-Term Memory (LSTM) networks \cite{graves2012long}, and Echo State Networks (ESN) \cite{jaeger2007echo} have shown superior performance. 

Hybrid approaches combine model-based and data-driven methods to improve RUL prediction. For instance, a hybrid model using the Double Exponential Degradation Model (DEDM) and Gated Recurrent Unit (GRU) network fused with a Bayesian neural network (BNN) offers enhanced predictions \cite{liang2024hybrid}. Despite their advantages, hybrid models still depend on early-cycle data, limiting their flexibility.


Accurately predicting the RUL of LIBs remains challenging due to the varying and volatile degradation of features across cycles, which necessitates a highly adaptable prediction model. The specific challenges will be discussed in the next section.



\subsection{Problem Statement}


As summarized in Table \ref{tab:rul_prediction_methods}, current state-of-the-art studies categorize RUL prediction methods into two primary approaches: historical data-independent methods, which estimate the current RUL based on present cycling data and a few preceding cycles, and historical data-dependent methods, which leverage early-cycle data to predict the battery’s full lifespan.

While historical data-dependent methods achieve reasonable accuracy in benchmark evaluations \cite{liang2024hybrid, wang2023remaining}, they are hindered by practical issues such as incomplete cycle data \cite{ma2024accurate}, unavailability of early-cycle records, vary operational conditions \cite{ibraheem2023capacity} throughout a battery's lifespan, and challenges in battery repurposing \cite{sanz2021remaining}. Therefore, historical data-independent approaches are more suitable for real-world scenarios.

\begin{table}[h]
    \centering
    \caption{Overview of RUL Prediction Methods in LIBs Research}
    \label{tab:rul_prediction_methods}
    \resizebox{\textwidth}{!}{
    \begin{tabular}{cccccc}
        \toprule
        \multicolumn{2}{c}{\textbf{Task}} &&&&  \\
        \addlinespace
        \cmidrule(lr){1-2}
        \addlinespace
         \multicolumn{1}{c}{\textbf{Method}} & \multicolumn{1}{c}{\textbf{Prediction targets}} & \multicolumn{1}{c}{\textbf{Reference}} & \multicolumn{1}{c}{\textbf{Dataset}} & \multicolumn{1}{c}{\textbf{Signal Preprocessing}} & \multicolumn{1}{c}{\textbf{Prediction Model}} \\
        \addlinespace
        \hline
        \addlinespace
               
        \multirow{6}{*}{Historical data-independent approach} & RUL (capacity) & \cite{ansari2023optimized} & NASA PCoE  \cite{NASA_dataset} & Feature extraction of temperature, & Cascaded forward neural network (CFNN)  \\
        & & & Toyota Research Institute (TRI)~\cite{severson2019data} & current and voltage &  \\
        & & & & Sliding window technique for denoising &\\
        \addlinespace
        
        & RUL (remaining cycles) & \cite{xia2023historical} & 2022 Li-Ion Health Prediction (LHP) \cite{ma2022real} & Feature-based condition extraction & Dual-input DNN \\  
        & & & & Sequential Feature Sampling &\\
        \addlinespace

        & RUL (remaining cycles) & \cite{ma2024accurate} & Oxford Battery \cite{zhu2022data} & Ageing-correlated parameter extraction & Physics-based DNN \\
        \addlinespace

        \hline
        \addlinespace
        
        \multirow{16}{*}{Historical data-dependent approach} & RUL (capacity) & \cite{qu2019neural} & 2016 NASA Battery ~\cite{zhou2016lithium} & CEEMDAN & Single-input PA-LSTM \\
        \addlinespace
        
        & RUL (capacity) & \cite{wang2023lithium} & 2007 NASA Battery ~\cite{saha2007battery} & Binary segmentation & Two-phase capacity degradation model \\
        & & & &  using particle filtering method &\\
        \addlinespace

        & RUL (capacity) & \cite{liu2024hybrid} & CALCE-CX2 and CALCE-CS2 \cite{vasan2013center} & EMD & CNN model predicting the maximum discharging capacity \\
        & & & & GRU-FC & \\
        \addlinespace
        
        & RUL (capacity) & \cite{wang2023remaining} & NASA PCoE  \cite{NASA_dataset} & Variational Modal Decomposition (VMD) & Bayesian optimization LSTM network \\
        & & & CALCE-CX2 and CALCE-CS2 \cite{vasan2013center} & Kullback-Leibler (KL) divergence & ESN \\
        \addlinespace

        & RUL (capacity) & \cite{liang2024hybrid} & NASA PCoE  \cite{NASA_dataset} & 
        Z-score normalization & GRU-CNN network \\
        & & & NASA Random Walk \cite{bole2014adaptation} & EMD & Double Exponential Degradation Model (DEDM) \\
        & & & & & BNN \\
        \addlinespace

        & Cycle life & \cite{severson2019data} & TRI \cite{severson2019data} & Feature extraction for the first 100 cycles &  Elastic net \\
        \addlinespace

        & Cycle life & \cite{ibraheem2023capacity} & TRI \cite{severson2019data} & 
        Statistical and gradient-based feature extraction & Extreme Gradient Boosting (XGBoost) \\
        \addlinespace
        
        & Cycle life, RUL (capacity) & \cite{ma2023two} & TRI \cite{severson2019data} & Spline function & CNN model \\
        & & & & (interpolates discharge capacities to length 1000) & Double exponential model (DEM) \\
        & & & & & Gaussian process regression (GPR) \\
        \addlinespace
        \bottomrule
    \end{tabular}
    }
\end{table}

Many existing methods predict RUL based on estimated maximum discharge capacity thresholds, representing the remaining life as a percentage. However, expressing the output as the number of remaining charge-discharge cycles provides clearer.

Small datasets, like the Oxford Battery dataset (13 cells) and NASA battery datasets (4–34 cells) limit model robustness in real-world failure prediction. In contrast, large datasets such as the TRI dataset (124 cells, fast-charging) and the LHP dataset (77 cells, diverse discharge) provide extensive charge-discharge scenarios, making them well-suited for both training and testing of data-independent models.


Signal preprocessing can be limited by high dimensionality, especially with variational decomposition methods like EMD and VMD, which preserve or expand the original signal size. In contrast, statistical feature extraction methods—such as mean and standard deviation—offer a low-dimensional, efficient alternative that captures essential characteristics, making them ideal for real-time and high-accuracy industrial applications.

Model-based and hybrid approaches typically rely on early-cycle data for RUL prediction, yet each battery exhibits unique degradation patterns over its lifespan requiring adaptive data-driven strategies. Moreover, Domain adaptation (DA) techniques such as domain-adversarial neural networks (DANN) \cite{ganin2016domain, ye2021state} and Generative Adversarial Networks (GANs) \cite{goodfellow2014generative} offer effective solutions for adapting to other source's degradation patterns to improve RUL prediction in the target domain. To address these challenges, our proposed approach will be represented in the next section.

\subsection{Main Contribution}  

The main contributions of this work are summarized as follows:

\begin{itemize}
    \item We propose a historical data-independent RUL prediction framework for lithium-ion batteries that relies solely on present and recent cycling data, eliminating the need for early-cycle information. This makes the approach suitable for real-time applications and scenarios with incomplete historical records.

    \item Our model integrates advanced deep learning components—including LSTM, Multihead Attention, and NODE blocks—as a powerful feature extractor, along with linear layers in the predictors, enabling the capture of both temporal and dynamic battery behaviors.

    \item We introduce a domain adaptation strategy that combines two predictors using trainable trade-off parameters and a Maximum Mean Discrepancy (MMD)-based loss to learn domain-invariant features, enhancing transferability from source to target domains.

    \item The framework includes a lightweight yet robust preprocessing pipeline-noise reduction, statistical feature extraction (e.g., mean, standard deviation), and normalization—to improve signal quality and reduce dimensionality for efficient, real-time prediction.

    \item Extensive evaluations on the two largest publicly available datasets of A123 APR18650M1A cells \cite{severson2019data, xia2023historical}, covering diverse charging and discharging conditions, validate the superior performance and practical applicability of our approach in real-world battery health management.
\end{itemize}

\section{Preliminaries}
This section presents an overview of the key components of the prediction model architecture, including LSTM, Multihead Attention, and Neural Ordinary Differential Equation (NODE) \cite{chen2018neural} blocks.

\subsection{LSTM \cite{graves2012long}:} \label{lstm}
A recurrent architecture designed to overcome the vanishing gradient problem by introducing gating mechanisms. Its operations are defined by:
\[
\begin{aligned}
i_t &= \sigma(W_i x_t + U_i h_{t-1} + b_i),\\[5pt]
f_t &= \sigma(W_f x_t + U_f h_{t-1} + b_f),\\[5pt]
o_t &= \sigma(W_o x_t + U_o h_{t-1} + b_o),\\[5pt]
\tilde{c}_t &= \tanh(W_c x_t + U_c h_{t-1} + b_c),\\[5pt]
c_t &= f_t \odot c_{t-1} + i_t \odot \tilde{c}_t,\\[5pt]
h_t &= o_t \odot \tanh(c_t),
\end{aligned}
\]
where \(x_t\) is the input at time \(t\), \(h_{t-1}\) is the previous hidden state, and \(\sigma\) and \(\tanh\) are the sigmoid and hyperbolic tangent activation functions, respectively.

\subsection{Multihead Attention \cite{vaswani2017attention}:} \label{multi_attention}
A critical mechanism in Transformer models, enabling the network to attend jointly to information from different subspaces. The basic building block is the scaled dot-product attention:
\[
\text{Attention}(Q, K, V) = \text{softmax}\left(\frac{QK^\top}{\sqrt{d_k}}\right)V,
\]
where \(Q\), \(K\), and \(V\) represent the query, key, and value matrices, respectively, and \(d_k\) is the dimensionality of the keys. In a multihead setting, multiple attention heads are computed:
\[
\text{head}_i = \text{Attention}(QW_i^Q, KW_i^K, VW_i^V),
\]
and their outputs are concatenated and linearly transformed:
\[
\text{MultiHead}(Q, K, V) = \text{Concat}(\text{head}_1, \ldots, \text{head}_h)W^O.
\]

\subsection{NODE \cite{chen2018neural}:} \label{node}
A framework that extends deep learning architecture by modeling continuous-time dynamics instead of discrete transformations between layers. In NODE, the evolution of a hidden state \(h(t)\) is governed by an ordinary differential equation (ODE):
\[
\frac{dh(t)}{dt} = f(h(t), t, \theta),
\]
where \(f\) is a neural network parameterized by \(\theta\). The final state \(h(t)\) is obtained by solving this ODE over a time interval, which provides a flexible and memory-efficient representation.


\section{Proposed method} 
\label{sec:proposed_method}
\subsection{Overall architecture}
\begin{figure}[H]
  \centering
  \includegraphics[width=\textwidth]{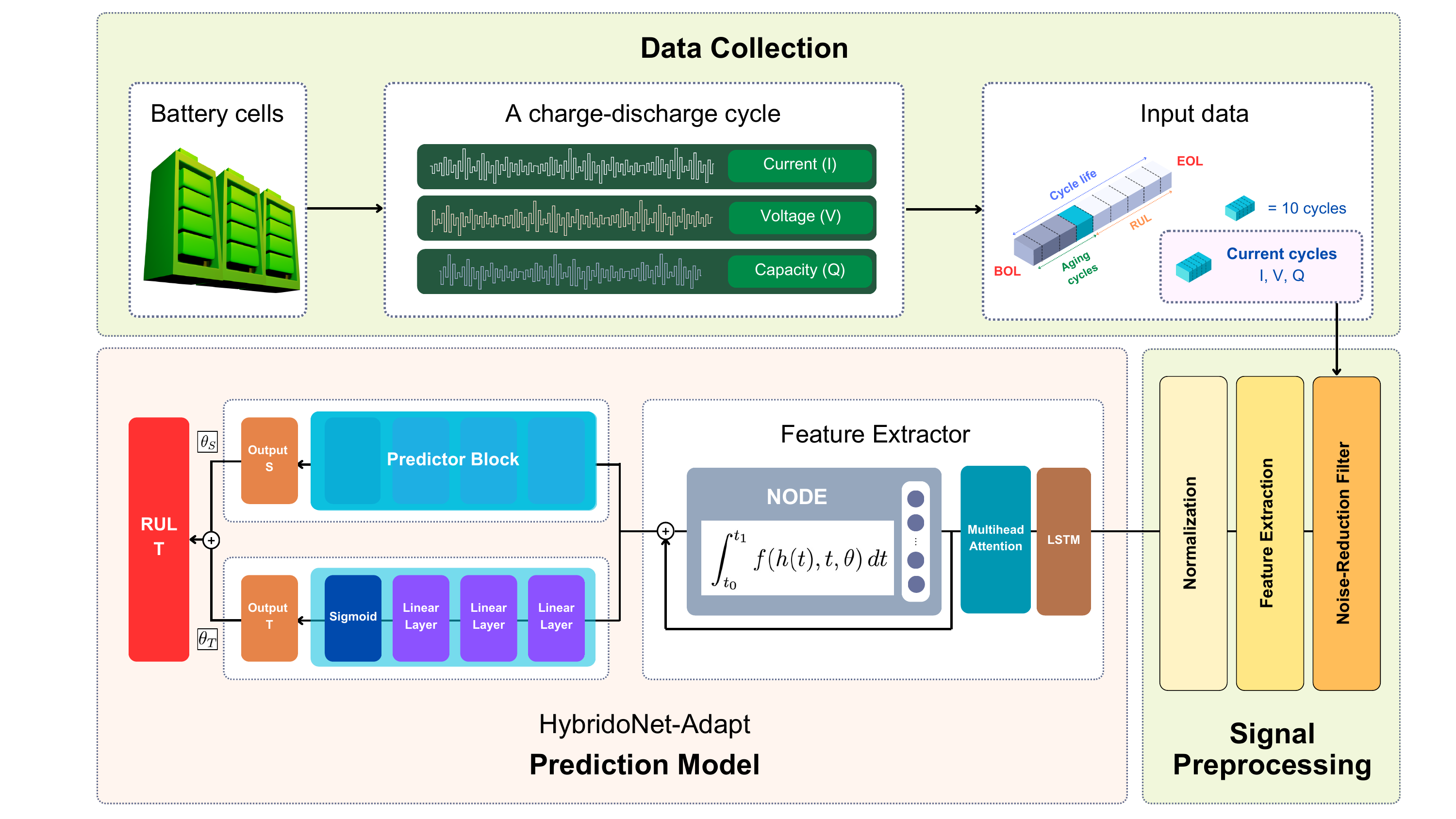}
  \caption{The overall RUL prediction process for Lithium-ion battery cells.}
  \label{fig:Overall_architecture}
\end{figure}

Figure \ref{fig:Overall_architecture} illustrates the RUL prediction process for Lithium-ion battery cells. In the data collection phase, Lithium Iron Phosphate (LFP)/graphite cells are monitored to capture voltage, current, and capacity signals for each individual charge-discharge cycle. 

Regarding cycle life degradation, the cycle life of a battery is defined as the total number of charge-discharge cycles from the Beginning of Life (BOL) to the End of Life (EOL). The EOL is typically identified when the battery's maximum capacity degrades to 70\% \cite{qu2019neural} or 80\% \cite{ma2023two} of its nominal capacity. The RUL, expressed in terms of the remaining number of cycles, is computed as:
\[
\text{RUL (remaining cycles)} = \text{Cycle life} - \text{Aging cycles},
\]
where \textit{Aging cycles} refers to the number of cycles the battery has already undergone.

In the signal preprocessing phase, the raw signals—voltage, current, and capacity from the most recent charge-discharge cycles—are passed through a noise-reduction filter named median filter \cite{justusson2006median} to smooth out sudden peaks. The filtered signals are then processed using feature extraction methods, including mean, standard deviation (Std), minimum (Min), maximum (Max), variance (Var), and median (Med) \cite{elangovan2011evaluation, sugumaran2011effect}. The extracted features for each cycle are represented as  
\[
X^{i} = \left[ x_{\text{current}}^{i}, x_{\text{voltage}}^{i}, x_{\text{capacity}}^{i} \right], \quad X^{i} \in \mathbb{R}^{3 \times 6},
\]
where the 3 rows correspond to the three signal types (voltage, current, and capacity), and the 6 columns represent the extracted features (Mean, Std, Min, Max, Var, Med). Each input sample to the prediction model consists of 10 selected cycles, uniformly sampled from a 30-cycle window (i.e., one cycle every three cycles) \cite{ma2022real}. The input sample is represented as  
\[
\mathbf{X^{i}} = \left[ X^{1}, X^{2}, \dots, X^{10} \right], \quad \mathbf{X} \in \mathbb{R}^{10 \times 3 \times 6}.
\]
Thus, the shape of the total target input data after the feature extraction step becomes  
\[
\mathbf{X^T} \in \mathbb{R}^{N \times 10 \times 3 \times 6}, 
\]
where \( N \) denotes the number of samples. During normalization, a MinMaxScaler is fitted and applied to scale each feature across all time steps and samples between 0 and 1. 

In the prediction phase, the RUL prediction model, named \textbf{HybridoNet-Adapt}, maps the target input \(\mathbf{X^T}\) to the predicted RUL \(\mathbf{Y^T} \in \mathbb{R}^{N \times 1}\). The details of the proposed RUL prediction model are presented in the following section.

\subsection{HybridoNet-Adapt: A Proposed RUL Prediction model with Novel Domain Adaptation}\label{subsec:RobustHybridoNet}
\begin{figure}[H]
  \centering
  \includegraphics[width=\textwidth]{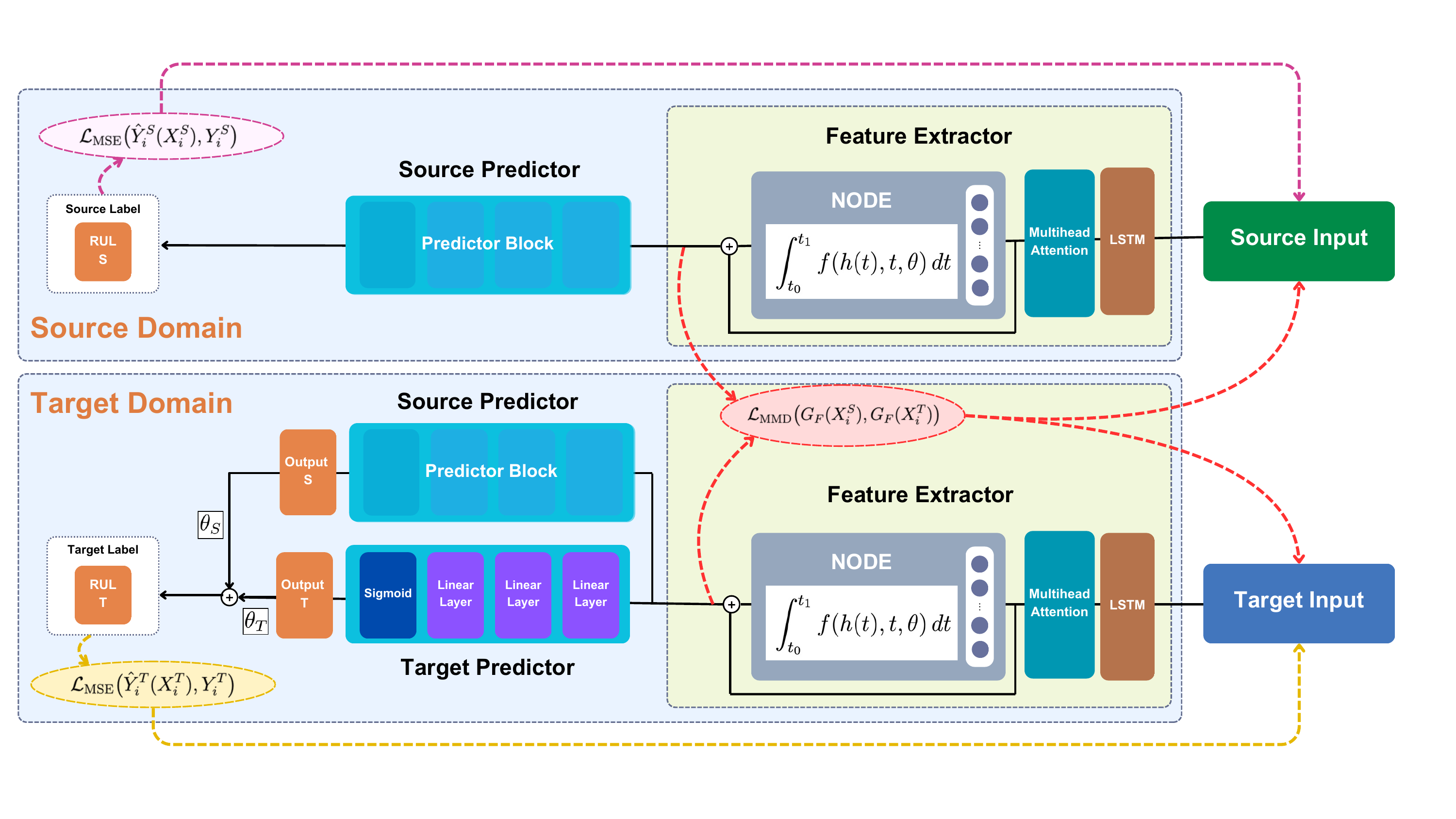}
  \caption{Architecture of the proposed HybridoNet-Adapt framework for RUL prediction with domain adaptation.}
  \label{fig:DA_training_process}
\end{figure}
As shown in Figure \ref{fig:DA_training_process}, HybridoNet-Adapt is composed of four key components: the source predictor \( G^{S}_{Y} \), the target predictor \( G^{T}_{Y} \), and the feature extractor \( G_{F} \), which is equipped with a DA technique to bridge the distribution gap between the source and target domains.

The feature extractor integrates a LSTM (Section \ref{lstm}), a Multihead Attention mechanism (Section \ref{multi_attention}), and a NODE block (Section \ref{node}). The NODE block models the hidden state \( h(t) \), which evolves continuously over time according to the following ODE:
\(
\frac{dh(t)}{dt} = f(h(t), t, \theta),
\)
where \( h(t) \) denotes the hidden state at time \( t \), \( f \) is a trainable function parameterized by \( \theta \), and \( t \) represents the continuous time variable. In our implementation, \( f \) is realized as a single linear layer to strike a balance between performance and computational efficiency. The initial condition for the NODE block is given by \( h(t_0) \), and the final transformed state \( h(t_1) \) is obtained by solving the ODE over the time interval \( [t_0, t_1] \):
\[
h(t_1) = h(t_0) + \int_{t_0}^{t_1} f(h(t), t, \theta) \, dt.
\]
In our experiments, the time bounds are set to \( t_0 = 0 \) and \( t_1 = 1 \), based on empirical results (see Figure~\ref{fig:time_steps_NODE}). The function \( h(t) \) thus represents the dynamic trajectory of the hidden state under continuous transformation, enabling the model to capture more nuanced temporal dependencies than discrete-time layers alone.

The source and target predictor modules share the same architecture, each consists of three linear layers, each linear layer followed by ReLU \cite{chen2020dynamic} activation, 1D Batch Normalization \cite{ioffe2015batch}, and Dropout \cite{hinton2012improving}, respectively, with a sigmoid activation \cite{yin2003flexible} layer at the end.

The target and source predictions in HybridoNet-Adapt are respectively computed as follows:
\begin{align}
\hat{Y}^{T}_{i}(X_{i}) &= \theta^{S} \, G^{S}_{Y}\big( G_{F}(X_{i}) \big) + \theta^{T} \, G^{T}_{Y}\big( G_{F}(X_{i}) \big), \\
\hat{Y}^{S}_{i}(X_{i}) &= G^{S}_{Y}\big( G_{F}(X_{i}) \big),
\end{align}
where \( \theta^{S} \) and \( \theta^{T} \) are learnable trade-off parameters that balance the contributions from the source and target predictors. The outputs \( \hat{Y}^{T}_{i} \) and \( \hat{Y}^{S}_{i} \) denote the target and source predictions, respectively.

To optimize the model, we employ a domain adaptation strategy that combines two loss functions: the mean squared error (MSE) \cite{das2004mean} loss, \( \mathcal{L}_{\text{MSE}} \), used for regression targets \( \hat{Y}^{T}_{i} \) and \( \hat{Y}^{S}_{i} \); and the maximum mean discrepancy (MMD) \cite{borgwardt2006integrating} loss, \( \mathcal{L}_{\text{MMD}} \), which encourages alignment between the feature distributions extracted from the source and target domains. The total loss function is defined as:
\begin{align*}
\mathcal{L}(X_{i}^{S}, X_{i}^{T}, Y_{i}^{S}, Y_{i}^{T}) =\ & 
\mathcal{L}_{\text{MSE}} \big( \hat{Y}^{S}_{i}(X^{S}_{i}), Y^{S}_{i} \big) \\
&+ \mathcal{L}_{\text{MSE}} \big( \hat{Y}^{T}_{i}(X^{T}_{i}), Y^{T}_{i} \big) \\
&+ \lambda \, \mathcal{L}_{\text{MMD}} \big( G_{F}(X^{S}_{i}), G_{F}(X^{T}_{i}) \big),
\label{eq:total_loss}
\end{align*}
where \( X^{S}_{i} \) and \( X^{T}_{i} \) are the input samples from the source and target domains, respectively, and \( Y^{S}_{i} \) and \( Y^{T}_{i} \) are their corresponding RUL labels. The hyperparameter \( \lambda \) controls the weight of the MMD loss in the overall objective.

The MMD loss quantifies the discrepancy between the distributions of source and target feature embeddings. Given extracted feature sets \( F^{S}_{i} \) from the source domain \( G_{F}(X^{S}_{i}) \) and \( F^{T}_{j} \) from the target domain \( G_{F}(X^{T}_{j}) \), it is defined as:  

\begin{align}
\mathcal{L}_{\text{MMD}}(F^{S}_{i}, F^{T}_{j}) &= \frac{1}{n^2} \sum_{i=1}^{n} \sum_{j=1}^{n} k(F^{S}_{i}, F^{S}_{j})  
+ \frac{1}{m^2} \sum_{i=1}^{m} \sum_{j=1}^{m} k(F^{T}_{i}, F^{T}_{j}) \notag \\
&\quad - \frac{2}{nm} \sum_{i=1}^{n} \sum_{j=1}^{m} k(F^{S}_{i}, F^{T}_{j}),
\end{align}
where \( k(\cdot, \cdot) \) is a kernel function, commonly chosen as the Gaussian kernel:
\(
k(x, y) = \exp\left(-\frac{\|x - y\|^2}{2\sigma^2}\right),
\)
with \(\sigma\) as the kernel bandwidth parameter. \(n\) and \(m\) denote the number of training samples from source and target domains, respectively.

The MSE loss is used to optimize the regression outputs by penalizing the squared differences between predicted and ground truth values. It is defined as:
\[
\mathcal{L}_{\text{MSE}}(\hat{Y_i}, Y_i) = \frac{1}{u} \sum_{i=1}^{u} \left( \hat{Y_i} - Y_i \right)^2,
\]
where \( \hat{Y}_i \) denotes the predicted value, and \( Y_i \) is the corresponding label for the RUL. \(u\) is the number of training samples.

The hidden dimension of both LSTM and Multihead Attention is set to 64, with 2 LSTM layers. The hidden dimensions of the linear layers in the source and target predictors are \([128, 64, 32, 1]\), with a dropout rate of 0.1. These values are determined based on the experimental results presented in Section \ref{HybridoNetAdapt-config}. 

In the following section, a series of experiments are conducted to identify the optimal configuration of HybridoNet-Adapt and to demonstrate its superiority over state-of-the-art methods. For comparison purposes, a supervised baseline model named \textit{HybridoNet} is constructed, consisting of the target predictor \( G^{T}_{Y} \) and the feature extractor \( G_{F} \). This model is trained using the MSE loss function. By comparing its performance with that of HybridoNet-Adapt, we highlight the performance improvements achieved through the incorporation of our proposed Domain Adaptation technique.

\section{Experiments and discussion} \label{sec:experimental_results}

\subsection{Experimental Setup} \label{subsec:experimental_setup}

Our proposed RUL model is implemented using the PyTorch framework and optimized using the AdamW algorithm \cite{loshchilov2017decoupled} to minimize the respective loss functions. All experiments are conducted on an NVIDIA A100 GPU with 80GB of memory. Each experiment is trained for 10 epochs with a batch size of 128 and a fixed learning rate of 0.0005. To reduce variability in the training process, each experiment is repeated 10 times, and the final prediction is computed as the average of these runs. The training data is divided into 90\% for training and 10\% for validation, with the model selected based on the lowest RMSE on the validation set (see Section~\ref{subsec:evaluation_metrics}). The weighting factor \(\lambda\) in Equation~\ref{eq:total_loss} is dynamically adjusted during training using the following schedule \cite{ganin2015unsupervised}:
\[
\lambda = \frac{2}{1 + e^{-10 \cdot \frac{\text{epoch}}{\text{epochs}}}} - 1,
\]
where \(\text{epoch}\) denotes the current training epoch, and \(\text{epochs}\) is the total number of training epochs.

\subsection{Datasets} 

\subsubsection{First dataset: Varied fast-charging conditions, with consistent discharging conditions}\label{subsec:firstDataset}

The first dataset, referred to as the TRI dataset \cite{severson2019data}, encompasses a detailed study of 124 LFP/graphite lithium-ion batteries. Each LIB in the dataset has a nominal capacity of 1.1 Ah and a nominal voltage of 3.3 V. The cycle life span of these batteries ranges from 150 to 2,300 cycles, showcasing a wide spectrum of longevity. In terms of operational conditions, all LIBs were subjected to uniform discharge protocols. Specifically, they were discharged at a constant current rate of 4 C until the voltage dropped to 2 V, followed by a constant voltage discharge at 2 V until the current diminished to C/50. The LIBs were charged at rates between 3.6 C and 6 C, under a controlled temperature of 30°C within an environmental chamber. The dataset contains approximately 96,700 cycles, making it one of the largest datasets to consider various fast-charging protocols. The dataset is divided into three distinct parts: a training set with 41 LIBs, a primary test set with 43 LIBs, and a secondary test set comprising 40 LIBs.

\subsubsection{Second dataset: Varied discharge conditions, with consistent fast-charging conditions}\label{subsec:secondDataset}

The second dataset, referred as the LHP dataset \cite{ma2022real}, was developed through a battery degradation experiment involving 77 cells (LFP/graphite A123 APR18650M1A) with a nominal capacity of 1.1 Ah and a nominal voltage of 3.3 V. Each of the 77 cells was subjected to a unique multi-stage discharge protocol, while maintaining an identical fast-charging protocol for all cells. The experiment was conducted in two thermostatic chambers at a controlled temperature of 30°C. The dataset encompasses a total of 146,122 discharge cycles, making it one of the largest datasets to consider various discharge protocols. The cells exhibit a cycle life ranging from 1,100 to 2,700 cycles, with an average of 1,898 cycles and a standard deviation of 387 cycles. The discharge capacity as a function of cycle number reveals a wide distribution of cycle lives. The dataset is divided into two distinct parts: a training set with 55 LIBs, and a test set with 22 LIBs.

\subsection{Evaluation Metrics} \label{subsec:evaluation_metrics}
To evaluate RUL prediction, we use Root Mean Square Error (\(RMSE\)) \cite{willmott2005advantages}, R-squared (\(R^2\)) \cite{cameron1997r, wang2023remaining}, and Mean Absolute Percentage Error (\(MAPE\)) \cite{de2016mean}. These are calculated as follows:

\[
RMSE(y_i, \hat{y}_i) = \sqrt{\frac{1}{n} \sum_{i=1}^{n} (y_i - \hat{y}_i)^2},
\]

\[
MAPE(y_i, \hat{y}_i) = \frac{1}{n} \sum_{i=1}^{n} \frac{|y_i - \hat{y}_i|}y \times 100,
\]

\[
R^2(y, \hat{y}) = 1 - \frac{\sum_{i=1}^{n} (y_i - \hat{y}_i)^2}{\sum_{i=1}^{n} (y_i - \bar{y})^2}.
\]
Where \( y_i \) and \( \hat{y}_i \) are the observed and predicted RUL, respectively. \(y\) is cycle life.The smaller the \(RMSE\) and \(MAPE\), and the larger the \(R^2\), the better the performance.




\subsection{Signal Analysis and Preprocessing}
\begin{figure}[H]
  \centering
  \begin{subfigure}{0.7\textwidth}
    \centering
    \includegraphics[width=\linewidth]{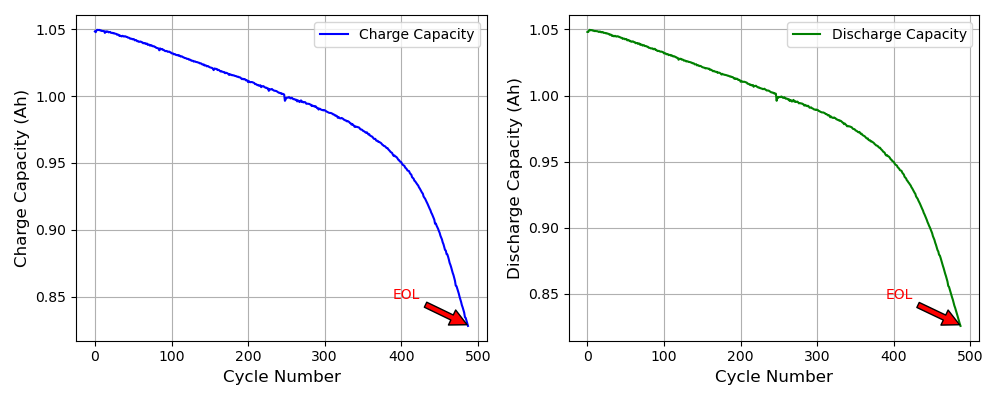}
    \caption{Maximum charge and discharge capacities over charge-discharge cycles for a single battery cell.}
    \label{fig:charge_discharge}
  \end{subfigure}
  
  \begin{subfigure}{0.7\textwidth}
    \centering
    \includegraphics[width=\linewidth]{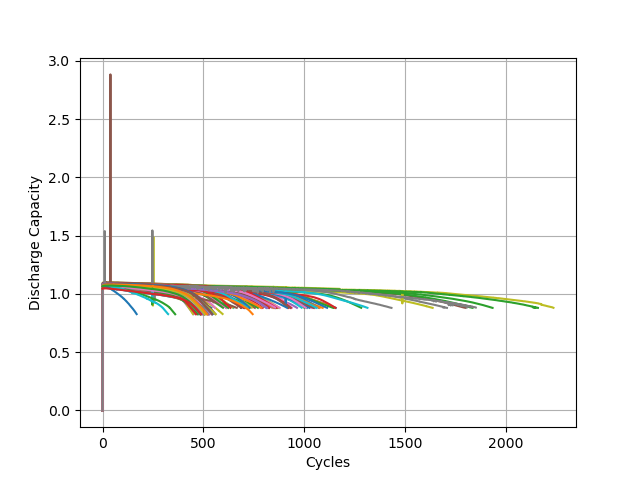}
    \caption{Maximum discharge capacities over charge-discharge cycles for many battery cells.}
    \label{fig:all_charge}
  \end{subfigure}
  
  \caption{Comparison of maximum charge and discharge capacities over cycle life for battery cells.}
  \label{fig:charge_comparison}
\end{figure}

Figures \ref{fig:charge_discharge} and \ref{fig:all_charge} analyze battery cycle life. Figure \ref{fig:charge_discharge} tracks an individual cell's charge and discharge capacities, marking EOL when the maximum capacity degrades to 80\% of nominal capacity. Figure \ref{fig:all_charge} compares cycle life across cells, revealing significant variation in discharge capacity. This variability challenges prediction models for RUL, emphasizing the need for accurate and adaptable RUL predictions for BHM systems.

\begin{figure}[h]
    \centering
    \resizebox{\textwidth}{!}{
    \includegraphics{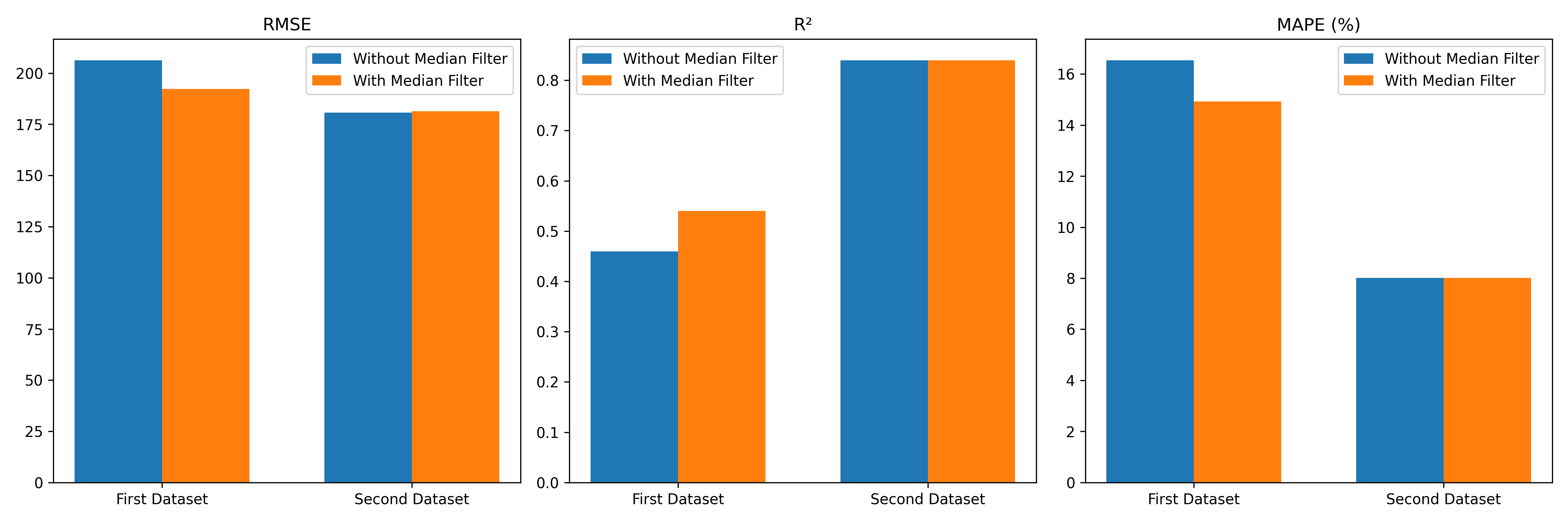}  
    }
    \caption{Comparison of RUL prediction performance with and without median filtering.}
    \label{fig:median_filter_comparison}
\end{figure}

Before feature extraction step in the signal preprocessing phase (as mentioned in \ref{sec:proposed_method}), the raw signals exhibit sudden peaks and fluctuations, resembling noise. Smoothing the time-series data can help reduce noise and enhance the key characteristics of the signal. To achieve this, a median filtering method is applied to eliminate abrupt peaks in the signals before feature extraction. As a result, the application of median filtering improves overall model performance. The filtered data leads to better RMSE, \(R^{2}\), and MAPE (\%) values compared to the unfiltered data, as illustrated in Figure \ref{fig:median_filter_comparison}.

\subsection{Feature Extractor}
\begin{figure}[h]
    \centering
    \includegraphics[width=0.7\textwidth]{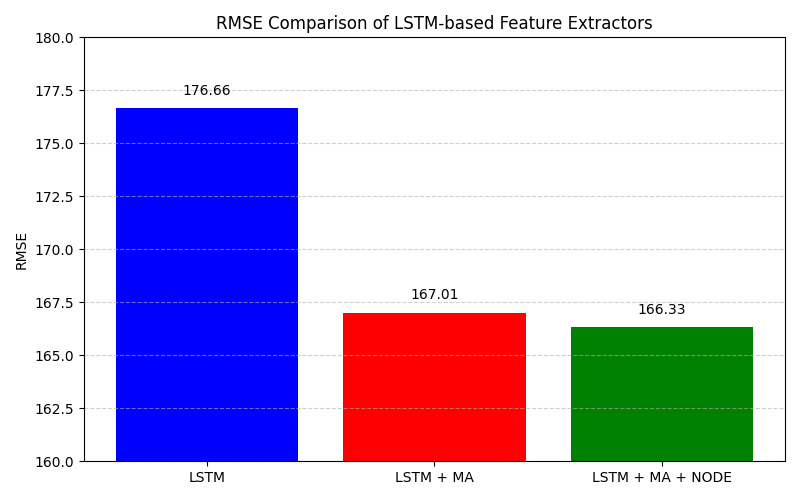}
    \caption{Performance comparison of LSTM-based blocks for feature extractor. Experiment on the testing data of the second dataset.}
    \label{fig:lstm_comparison}
\end{figure}

The feature extractor is progressively developed, starting with an LSTM architecture and sequentially integrating Multihead Attention (MA) and a NODE block. To evaluate the effectiveness of each component, we assess the performance of HybridoNet-Adapt at different stages. With each addition as shown in Figure \ref{fig:lstm_comparison}, the model's predictive capability improves. Ultimately, HybridoNet-Adapt achieves an RMSE of 166.33, an \( R^{2} \) score of 0.86, and a MAPE of 7.44\%, demonstrating its superior performance.


\subsection{HybridoNet-Adapt with Domain Adaption}\label{HybridoNetAdapt-config}
\begin{figure}[h]
    \centering
    \includegraphics[width=1\textwidth]{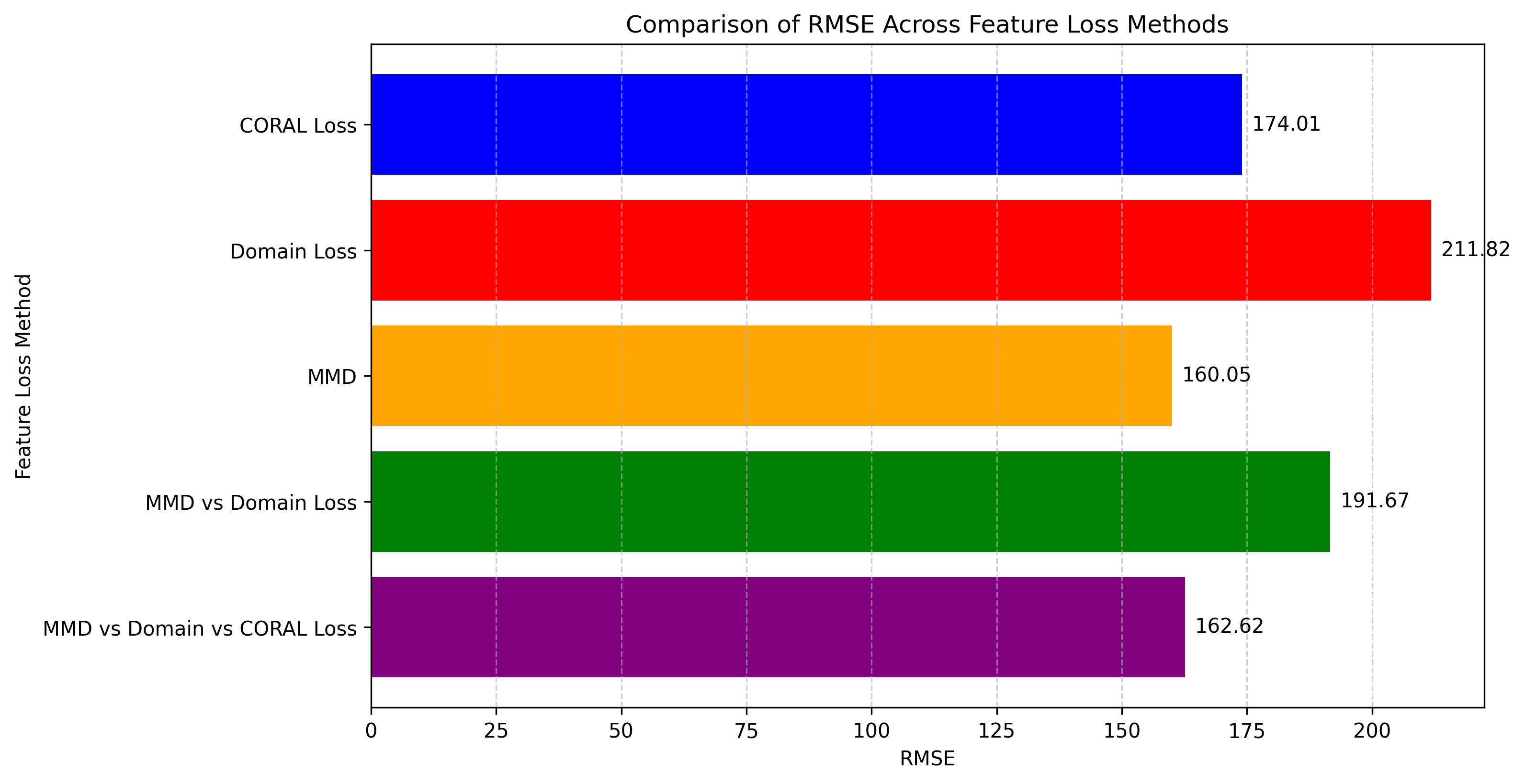}
    \caption{Comparison of different feature loss methods. Experiment on the testing data of the second dataset.}
    \label{fig:feature_loss_comparison}
\end{figure}

HybridoNet-Adapt is evaluated with various feature loss functions, including CORAL Loss, Domain Loss \cite{ganin2016domain}, MMD, as well as combinations such as MMD with Domain Loss, and MMD with Domain Loss and CORAL Loss, as shown in Figure \ref{fig:feature_loss_comparison}. The results indicate that using only MMD as the feature loss function yields the best performance, achieving an RMSE of 160.05.

\begin{figure}[H]
    \centering
    \includegraphics[width=1\textwidth]{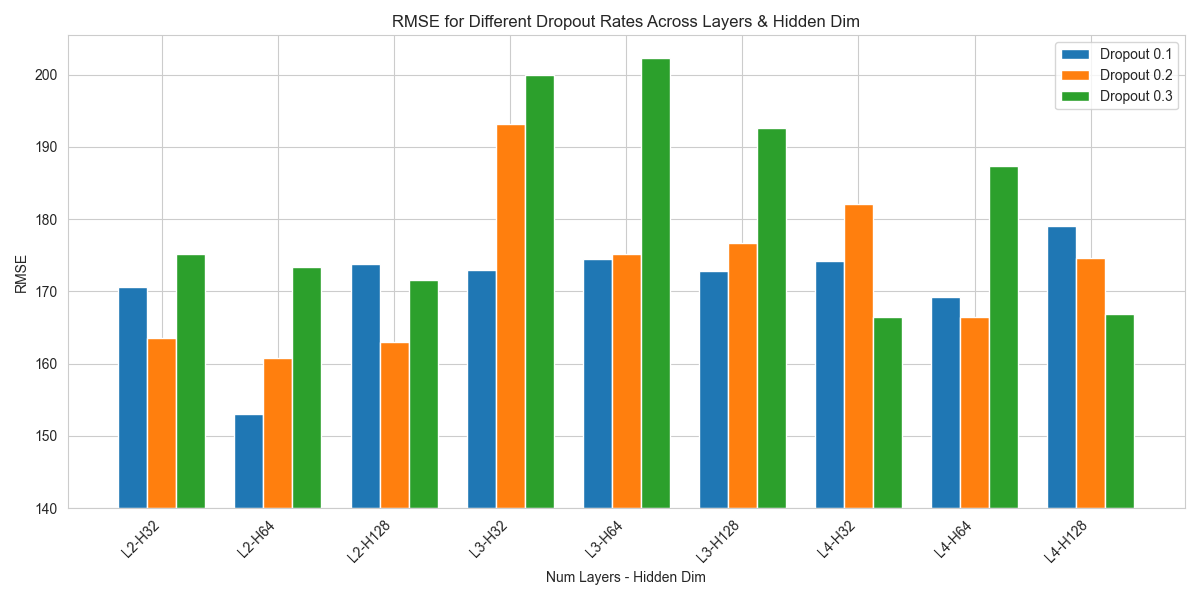}
    \caption{Comparison of RMSE across different number of LSTM layers, hidden dimensions, and dropout configurations. Experiment on the testing data of the second dataset.}
    \label{fig:Search_results_num_layers_hidden_dim_dropout}
\end{figure}

To determine the optimal hyperparameters, including hidden dimension of all layers, the number of recurrent LSTM layers, and the dropout rate, 27 experiments were conducted. The results are presented in Figure \ref{fig:Search_results_num_layers_hidden_dim_dropout}. In the graph, \(L\) represents the number of recurrent layers, \(H\) denotes the hidden dimension size. Based on RMSE score, the best performance is achieved with 2 recurrent LSTM layers, a hidden dimension of 64, and a dropout rate of 0.1. 

\begin{figure}[H]
    \centering
    \includegraphics[width=0.7\textwidth]{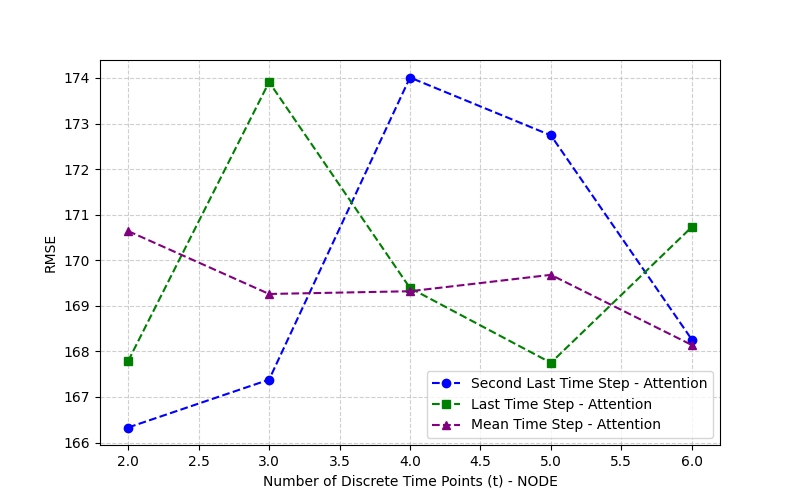}
    \caption{Comparison of RMSE for different NODE discrete time steps ($t$) and Multihead Attention output time step selections. Experiment on the testing data of the second dataset.}
    \label{fig:time_steps_NODE}
\end{figure}

To identify the optimal time step in the sequence dimension for both Multihead Attention and NODE outputs, a comprehensive evaluation was conducted. Various NODE output time steps ranging from 2 to 6 were tested, along with different Multihead Attention output time step selections, including the last, the second-to-last, and the mean time step. As shown in Figure \ref{fig:time_steps_NODE}, the best performance was achieved when using the second-to-last time step of the Multihead Attention output and a NODE output time step of 2.

The proposed HybridoNet-Adapt model is systematically evaluated under various scenarios by experimenting with four different target sets, each derived from the training data of the second dataset. The source data is the training data from the first dataset. Below are four target groups of battery cells selected from the training data of the second dataset. These groups are carefully formed to ensure each set represents a diverse range of battery performance. For instance, Group 1 includes both high-cycle cells (e.g., 2-2 with 2,651 cycles) and low-cycle cells (e.g., 1-6 with 1,143 cycles), ensuring a comprehensive representation of aging behaviors.

\begin{itemize}
    \item \textbf{Group 1:} 1-3 (1,858 cycles), 1-6 (1,143 cycles), 2-2 (2,651 cycles), 2-6 (1,572 cycles), 3-2 (2,283 cycles), 3-6 (2,491 cycles), 4-3 (1,142 cycles), 5-4 (1,962 cycles)
    \item \textbf{Group 2:} 1-5 (1,971 cycles), 1-8 (2,285 cycles), 2-4 (1,499 cycles), 2-7 (2,202 cycles), 3-3 (1,649 cycles), 3-7 (2,479 cycles), 4-4 (1,491 cycles), 5-5 (1,583 cycles)
    \item \textbf{Group 3:} 2-8 (1,481 cycles), 3-4 (1,766 cycles), 3-8 (2,342 cycles), 4-1 (2,217 cycles), 4-7 (2,216 cycles), 5-1 (2,507 cycles), 5-6 (2,460 cycles), 6-3 (1,804 cycles)
    \item \textbf{Group 4:} 4-8 (1,706 cycles), 5-2 (1,926 cycles), 5-7 (1,448 cycles), 6-4 (1,717 cycles), 6-5 (2,178 cycles), 7-2 (2,030 cycles), 7-7 (1,685 cycles), 8-2 (2,041 cycles)
    \item \textbf{All:} All battery cells from the training set of the second dataset.
\end{itemize}
Table \ref{tab:DA_Comparison} shows that HybridoNet-Adapt outperforms both HybridoNet (without DA) and DANN (with DA) across all groups. It achieves the lowest RMSE and MAPE while maintaining the highest \(R^2\), demonstrating better generalization. For instance, in Group 1, HybridoNet-Adapt reduces RMSE from 368.99 to 356.46 and improves \(R^2\) from 0.21 to 0.30. On the full dataset, it achieves the best RMSE of 153.24 and \(R^2\) of 0.88, significantly outperforming DANN, which shows degraded performance (RMSE = 835.35, \(R^2 = -1.37\)). DANN struggles with large variations in battery aging, while HybridoNet-Adapt effectively adapts to different distributions, leading to consistently better predictions.

\begin{table}[h]
    \centering
    \caption{Comparison of HybridoNet, DANN, and HybridoNet-Adapt across four target data groups from the testing data of the second dataset.} 
    \label{tab:DA_Comparison}
    \resizebox{\textwidth}{!}{
    \begin{tabular}{lccccccccc}
        \toprule
        \multirow{2}{*}{\textbf{Group}} & \multicolumn{3}{c}{\textbf{HybridoNet (Without DA)}} & \multicolumn{3}{c}{\textbf{DANN (With DA)}} & \multicolumn{3}{c}{\textbf{HybridoNet-Adapt (With DA)}} \\
        \cmidrule(lr){2-4} \cmidrule(lr){5-7} \cmidrule(lr){8-10}
        &  \textbf{RMSE $\downarrow$} & \textbf{R$^2$ $\uparrow$} & \textbf{MAPE (\%) $\downarrow$} & \textbf{RMSE $\downarrow$} & \textbf{R$^2$ $\uparrow$} & \textbf{MAPE (\%) $\downarrow$} & \textbf{RMSE $\downarrow$} & \textbf{R$^2$ $\uparrow$} & \textbf{MAPE (\%) $\downarrow$} \\
        \midrule
        Group 1  & 368.99 & 0.21 & 18.26 & 604.14 & -0.36 & 27.26 & 356.46 & 0.30 & 17.35 \\
        Group 2  & 245.58 & 0.71 & 11.84 & 665.02 & -0.51 & 29.33 & 240.90 & 0.73 & 11.09 \\
        Group 3  & 334.79 & 0.35 & 16.69 & 1007.93 & -3.39 & 51.02 & 316.79 & 0.41 & 15.43 \\
        Group 4  & 304.91 & 0.61 & 14.08 & 758.16 & -0.99 & 32.84 & 293.40 & 0.63 & 13.27 \\
        \midrule
        All      & 166.33 & 0.86 & 7.44  & 835,35    & -1,37  & 36,99   & 153.24 & 0.88 & 7.30 \\
        \bottomrule
    \end{tabular}
    }
\end{table}

Table \ref{tab:RUL_result_seconddataset} presents the evaluation metrics for RUL prediction on the test data from the second dataset, comparing Elastic Net, $A_1$, $A_2$ of paper \cite{ma2022real} (see Table S4), with our HybridoNet, and HybridoNet-Adapt methods. The results indicate that HybridoNet-Adapt achieves competitive RMSE values, particularly in cases where Elastic Net exhibits high errors. The $R^2$ values show that HybridoNet-Adapt generally improves predictive accuracy compared to the baseline methods. Additionally, MAPE results suggest that HybridoNet-Adapt provides more stable and reliable predictions, especially in challenging scenarios. Overall, these findings demonstrate the potential of HybridoNet-Adapt for enhanced RUL estimation.

\begin{table}[H]
    \centering
    \caption{Evaluation metrics for RUL prediction performance  using existing Elastic Net, \(A_{1}\), and \(A_{2}\) results of paper \cite{ma2022real} (see Table S4). Experiment on the testing data of the second dataset.}  
    \label{tab:RUL_result_seconddataset}
    \resizebox{\textwidth}{!}{
        \begin{tabular}{ccccccccccccccccc}
            \toprule
            & & \multicolumn{5}{c}{\textbf{RMSE (cycles)}} & \multicolumn{5}{c}{$\mathbf{R^2}$} & \multicolumn{5}{c}{\textbf{MAPE (\%)}} \\
            
            \cmidrule(lr){3-7} \cmidrule(lr){8-12} \cmidrule(lr){13-17} 
            
            \textbf{Protocol} & \textbf{Channel} & \textbf{Elastic net} & $\textbf{A}_{1}$ & $\textbf{A}_{2}$ & \textbf{HybridoNet} & \textbf{\text{HybridoNet-Adapt}} & \textbf{Elastic net} & $\textbf{A}_{1}$ & $\textbf{A}_{2}$ & \textbf{HybridoNet} & \textbf{\text{HybridoNet-Adapt}} & \textbf{Elastic net} & $\textbf{A}_{1}$ & $\textbf{A}_{2}$ & \textbf{HybridoNet} & \textbf{\text{HybridoNet-Adapt}} \\
            \midrule
            \#1 & 1-1   & 252 & 63.8 & 42.7 & 30,9 & 57,84 & 0.646  & 0.977 & 0.990 & 0,99 & 0,98 & 14.1 & 3.64 & 2.16 & 1,67 & 3,39  \\
            \#2 & 1-2   & 722 & 262  & 272  & 483,29 & 514,39 & 0.102  & 0.882 & 0.873 & 0,6 & 0,55 & 23.2 & 8.64 & 8.52 & 15,22 & 17,34 \\
            \#12 & 2-5  & 365 & 390  & 364  & 96,9 & 158,42 & 0.122  & /  & 0.126 & 0,94 & 0,84 & 19.4 & 23.5 & 22.0 &  5,46 & 9,38 \\
            \#16 & 3-1  & 422 & 104  & 133  & 327,53 & 129,66 & 0.407  & 0.964 & 0.941 & 0,65 & 0,94 & 17.0 & 4.43 & 6.00 & 15,39 &  6,42 \\
            \#28 & 4-5  & 313 & 307  & 301  & 62,55 & 125,98 & 0.493  & 0.512 & 0.531 & 0,98 & 0,92 & 16.1 & 15.7 & 16.3 & 2,69 & 7,4 \\
            \#34 & 5-3  & 757 & 279  & 301  & 346,43 & 392,7 & 0.0225 & 0.867 & 0.845 & 0,8 & 0,74 & 24.2 & 8.94 & 9.47 & 11,18 & 12,4  \\
            \#39 & 6-1  & 310 & 147  & 120  & 70,95 & 34,57 & 0.532  & 0.896 & 0.930 & 0,98 & 0,99 & 14.3 & 7.69 & 6.01 & 3,36 & 1,69 \\
            \#40 & 6-2  & 413 & 96.3 & 141  & 140,41 & 104,43 & 0.415  & 0.968 & 0.932 & 0,93 & 0,96 & 15.5 & 4.08 & 5.98 & 6,37 & 5,03 \\
            \#44 & 6-6  & 672 & 400  & 349  & 226,53 & 248,97 & 0.0821 & 0.675 & 0.753 & 0,9 & 0,87 & 22.4 & 14.5 & 12.6 & 8,71 & 9,52 \\
            \#45 & 6-8  & 609 & 334  & 359  & 148,63 & 217,09 & 0.236  & 0.769 & 0.734 & 0,95 & 0,9 & 20.3 & 12.6 & 13.6 & 5,23 & 8,3 \\
            \#50 & 7-5  & 372 & 58.0 & 46.2 & 68,29 & 142,18 & 0.508  & 0.988 & 0.992 & 0,98 & 0,93 & 15.6 & 2.70 & 2.01 & 2,53 & 6,52 \\
            \#51 & 7-6  & 372 & 295  & 263  & 50,22 & 81,06 & 0.129  & 0.453 & 0.566 & 0,98 & 0,96 & 21.5 & 15.7 & 15.6 & 2,65 & 5 \\
            \#54 & 8-1  & 303 & 200  & 231  & 331,1 & 319,56 & 0.319  & 0.702 & 0.603 & 0,19 & 0,25 & 19.7 & 12.8 & 15.6 & 23,66 & 23,94 \\
            \#58 & 8-5  & 281 & 45.3 & 57.1 & 229,23 & 227,22 & 0.449  & 0.986 & 0.977 & 0,64 & 0,64 & 17.7 & 2.64 & 3.30 & 15,19  & 16,5 \\
            \#59 & 8-6  & 527 & 91.8 & 81.5 & 45,05 & 116,28 & 0.386  & 0.981 & 0.985 & 1 & 0,97 & 18.1 & 3.41 & 2.97 & 1,7 & 4,29 \\
            \#61 & 8-8  & 412 & 363  & 382  & 254,98 & 49,44 & 0.245  & 0.411 & 0.349 & 0,71 & 0,99 & 18.2 & 18.2 & 19.3 & 13,3 & 2,31 \\
            \#65 & 9-4  & 431 & 104  & 74.3 & 102,13 & 34,23 & 0.406  & 0.966 & 0.982 & 0,97 & 1 & 16.9 & 4.56 & 3.17 & 3,79 & 1,43 \\
            \#67 & 9-6  & 403 & 297  & 292  & 182,32 & 56,22 & 0.328  & 0.634 & 0.649 & 0,86 & 0,99 & 18.4 & 15.4 & 15.1 & 6,24  & 2,97 \\
            \#70 & 10-1 & 386 & 111  & 81.0 & 27,28 & 57,33 & 0.355  & 0.947 & 0.972 & 1 & 0,99 & 17.6 & 5.03 & 4.24 & 1,28 & 2,52 \\
            \#73 & 10-4 & 331 & 67.5 & 60.9 & 184,35 & 43,13 & 0.582  & 0.983 & 0.986 & 0,87 & 0,99 & 14.3 & 3.15 & 2.95 & 8,36 & 2,05 \\
            \#75 & 10-6 & 485 & 112  & 84.0 & 188,2 & 103,52 & 0.442  & 0.970 & 0.983 & 0,92 & 0,97 & 16.0 & 3.93 & 3.12 & 6,62 & 3,96 \\
            \#76 & 10-7 & 405 & 86.7 & 52.8 & 62,06 & 157,16 & 0.353  & 0.970 & 0.989 & 0,98 & 0,9 & 16.7 & 3.25 & 1.97 & 3,13 & 8,28 \\
            \midrule
            \textbf{Mean} & - & 434 & 192 & 186 & 166,33 & 153,24 & 0.344 & 0.795 & 0.804 & 0,86 & 0,88 & 18.1 & 8.84 & 8.72 & 7,44 &  7,3\\
            \bottomrule
        \end{tabular}
    }
    
\end{table}

Table~\ref{tab:first_dataset} presents a performance comparison of different models on the secondary test data from the first dataset. XGBoost and HybridoNet are trained using the training data from the first dataset. HybridoNet-Adapt is trained with the second dataset's training data as the source input, and the first dataset's training data as the target input. HybridoNet-Adapt achieves the best results, with the lowest RMSE (146.52), the highest \( R^2 \) score (0.72), and the lowest MAPE (11.85\%), demonstrating its superior predictive accuracy with domain adaptation. Additionally, HybridoNet outperforms XGBoost in all metrics, highlighting the effectiveness of deep learning-based approaches. The improvements seen in HybridoNet-Adapt further validate the benefits of domain adaptation in enhancing RUL prediction performance.

\begin{table}[h]
    \centering
    \caption{Evaluation metrics for RUL prediction performance. Experiment on the secondary testing data of the first dataset.}
    \label{tab:first_dataset}
    \begin{tabular}{lccc}
        \toprule
        \textbf{Model} & \textbf{RMSE $\downarrow$} & \textbf{R² $\uparrow$} & \textbf{MAPE (\%) $\downarrow$} \\
        \midrule
        XGBoost  & 192.28 & 0.54  & 14.92  \\
        HybridoNet & 153.47 & 0.68 & 12.71\\
        HybridoNet-Adapt & 146.52 & 0.72 & 11.85 \\
        \bottomrule
    \end{tabular}
\end{table}

\begin{figure}[H]
    \centering
    \begin{minipage}{0.48\textwidth}
        \centering
        \includegraphics[width=\linewidth]{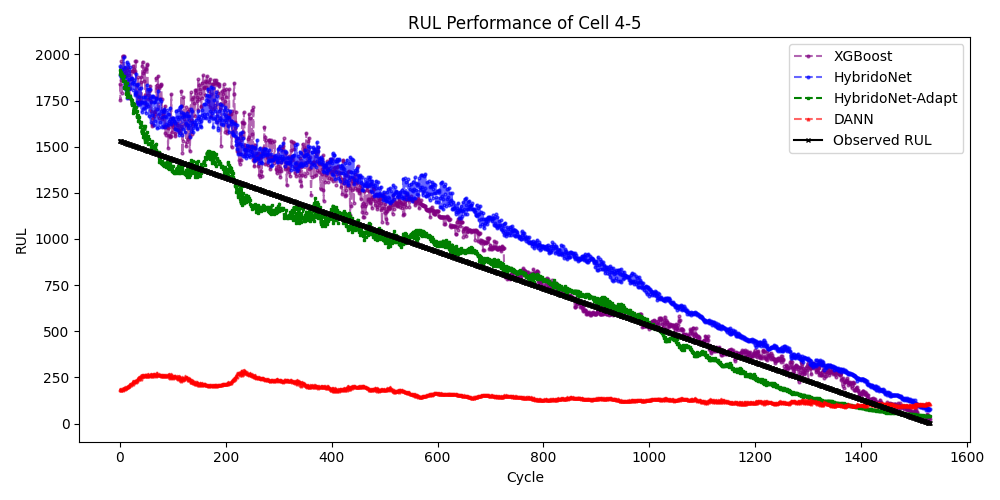}
    \end{minipage}
    \hfill
    \begin{minipage}{0.48\textwidth}
        \centering
        \includegraphics[width=\linewidth]{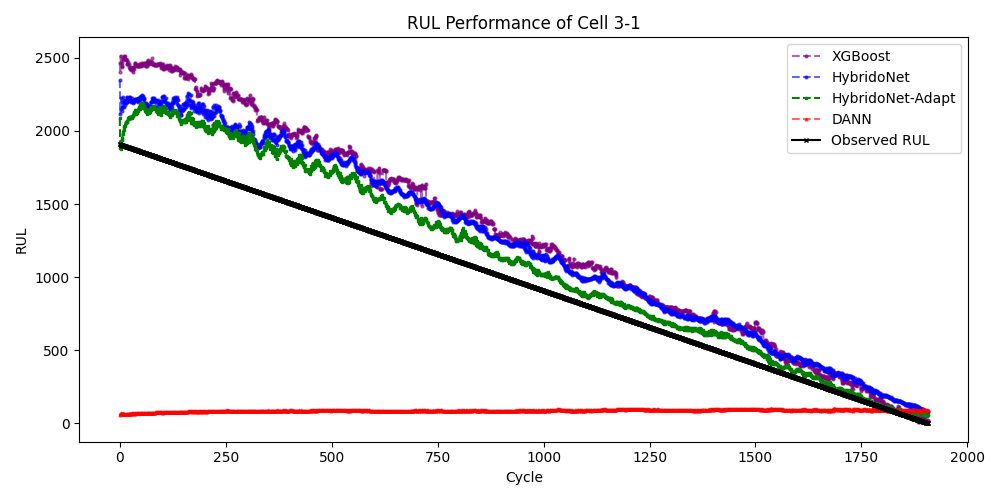}
    \end{minipage}
    \caption{Comparison of model predictions with observed RUL for Cells 4-5 and 3-1 from the testing data of the second dataset.}
    \label{fig:cell_comparison}
\end{figure}

\begin{figure}[H]
    \centering
    \includegraphics[width=\textwidth, keepaspectratio]{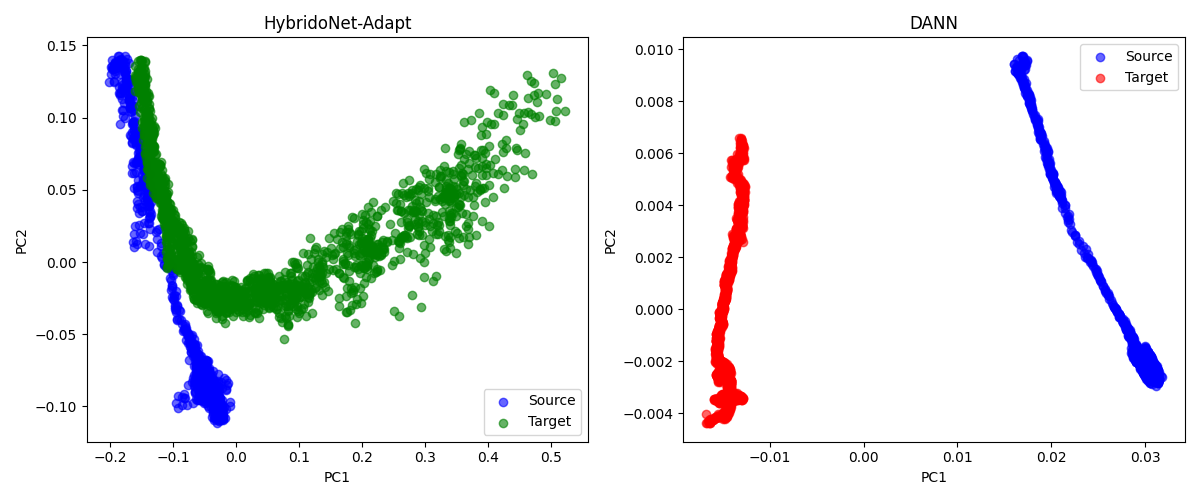}
    \caption{PCA-based comparison of embedding features between HybridoNet-Adapt and DANN from Cell 3-1 the testing data of the second dataset.}
    \label{fig:embedding_comparison}
\end{figure}

Figure \ref{fig:cell_comparison} illustrates the RUL predictions of XGBoost, HybridoNet, HybridoNet-Adapt, and DANN, compared to the true (observed) RUL for Cell 4-5 and Cell 3-1 in the testing set of the second dataset. Among all methods, HybridoNet-Adapt demonstrates the closest alignment with the observed RUL, highlighting its superior predictive accuracy. This improvement is attributed to HybridoNet-Adapt's ability to align feature representations from the source domain to the target domain, as shown in Figure \ref{fig:embedding_comparison}. By effectively increasing the amount of target-relevant data through our domain adaptation technique, HybridoNet-Adapt enhances robustness, making it more adaptable to diverse real-world battery degradation scenarios.

\subsection{Comparison with state-of-the-art methods}

\begin{figure}[h]
    \centering
    \includegraphics[width=1\textwidth]{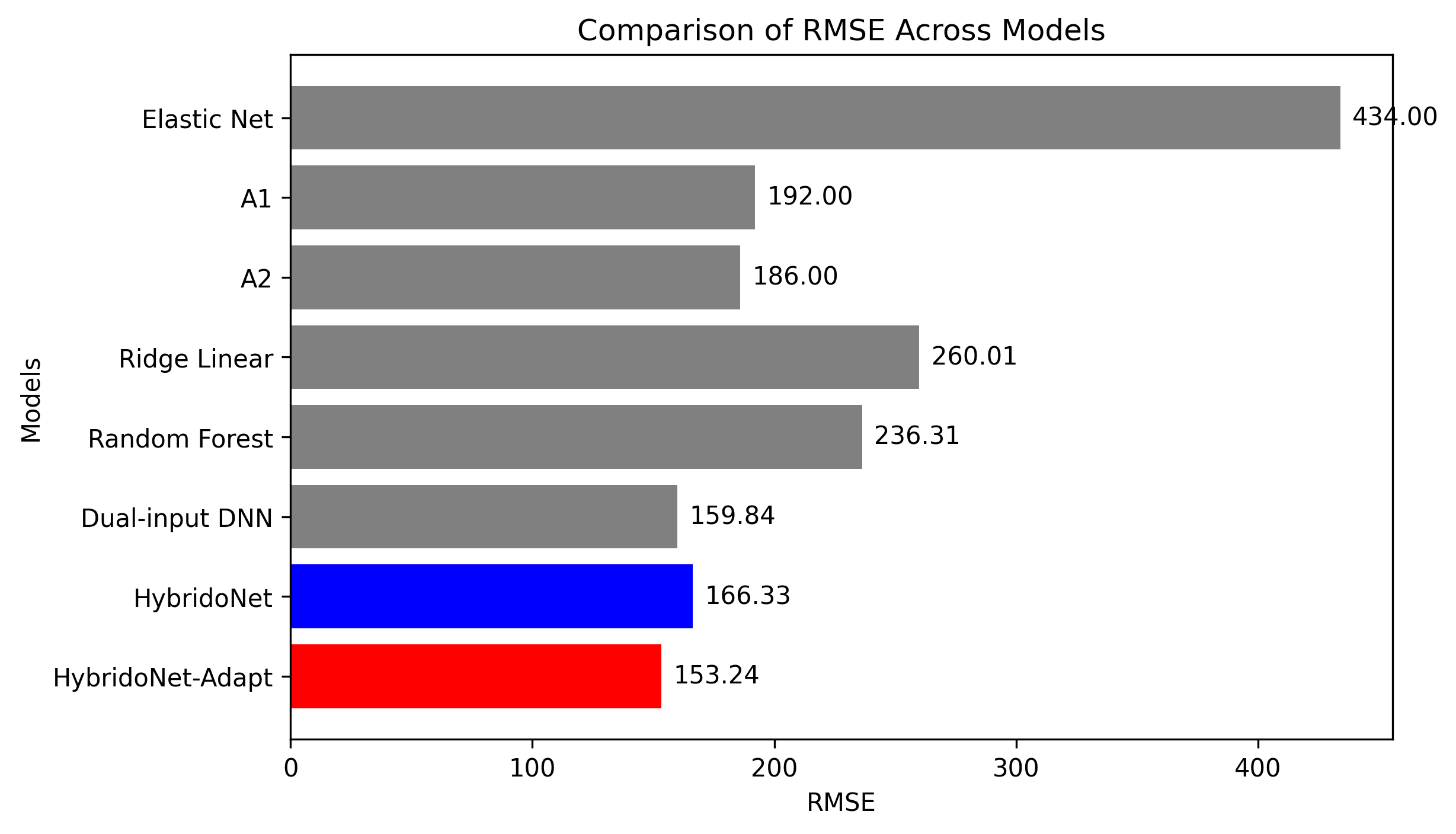}
    \caption{Comparison of our proposed models with existing state-of-the-art methods. Experiment on the testing data of the second dataset..}
    \label{fig:hybridoNet_comparison}
\end{figure}

Figure \ref{fig:hybridoNet_comparison} presents a comparison of our HybridoNet and HybridoNet-Adapt models with state-of-the-art methods, including Elastic Net \cite{ma2022real}, $\text{A}_{1}$ \cite{ma2022real}, $\text{A}_{2}$ \cite{ma2022real}, Ridge Linear \cite{xia2023historical}, Random Forest \cite{xia2023historical}, and Dual-input DNN \cite{xia2023historical}. The results demonstrate that our HybridoNet-Adapt achieves the lowest RMSE of 153.24, outperforming all other approaches, particularly the Dual-input DNN \cite{xia2023historical}, which attains an RMSE of 159.84. This highlights the effectiveness of our proposed method in enhancing predictive performance.

\section{Conclusion} \label{sec:conclusion}
In this study, we proposed a Remaining Useful Life (RUL) prediction framework comprising a signal processing module and a novel prediction model trained using a domain adaptation technique. In the signal preprocessing phase, raw signals undergo a noise-reduction filter, followed by feature extraction and normalization to prepare the data for prediction model input. The proposed prediction model, \textit{HybridoNet-Adapt}, consists of a feature extraction block and two predictors, which are constructed using LSTM layers, Multihead Attention mechanisms, Neural Ordinary Differential Equation (NODE) blocks, and linear layers. The outputs of the two predictors are balanced using two trainable trade-off parameters. To train the model, we introduce a domain adaptation strategy that combines Mean Squared Error (MSE) with Maximum Mean Discrepancy (MMD), enabling the model to learn domain-invariant features from the source domain and effectively generalize to the target domain. Experimental results demonstrate that \textit{HybridoNet-Adapt} outperforms conventional machine learning models such as XGBoost and Elastic Net, as well as state-of-the-art deep learning models like Dual-input DNN, affirming its effectiveness in the RUL prediction task. For future work, we plan to enhance model generalization through self-supervised learning, optimize strategies for real-time deployment, and explore multi-modal data integration to improve scalability and robustness across diverse battery health management (BHM)  applications.


\section*{Preprint Availability}
A preprint of this manuscript is available at:
\url{https://arxiv.org/pdf/2503.21392}

\section*{Contact Information}
For access to the code and further information about these proposed systems, please contact AIWARE Limited Company at: \url{https://aiware.website/Contact}

\bibliographystyle{plain}
\bibliography{cas-refs}
\end{document}